\documentclass{article}

\usepackage[preprint]{neurips_2024}

\usepackage[utf8]{inputenc} %
\usepackage[T1]{fontenc}    %
\usepackage{hyperref}       %
\usepackage{url}            %
\usepackage{booktabs}       %
\usepackage{amsfonts}       %
\usepackage{nicefrac}       %
\usepackage{microtype}      %
\usepackage{xcolor}         %
\usepackage[pdftex]{graphicx}
\usepackage[font=small,labelfont=bf]{caption}
\usepackage{subcaption}
\usepackage{float}
\usepackage{wrapfig}
\title{Inference-Friendly Models With MixAttention}
\usepackage{cleveref}

\author{%
  Shashank Rajput \\
  Databricks Mosaic AI\\
  \texttt{shashank.rajput@databricks.com} \\
  \And
  Ying Sheng \\
  Stanford University\thanks{Work done while at Databricks Mosaic AI.}\\
  \texttt{sqy1415@gmail.com} \\
    \And
Sean Owen \\
  Databricks Mosaic AI\\
  \texttt{sean.owen@databricks.com} \\
  \And
  Vitaliy Chiley \\
  Databricks Mosaic AI\\
  \texttt{vitaliy.chiley@databricks.com} \\
}

\usepackage{soul}

\begin{document}

\maketitle

\begin{abstract}

   The size of the key-value (KV) cache plays a critical role in determining both the maximum context length and the number of concurrent requests supported during inference in modern language models. The KV cache size grows proportionally with the number of attention heads and the tokens processed, leading to increased memory consumption and slower inference for long inputs. In this work, we explore the use of MixAttention, a model architecture modification closely related to a blog published by Character.AI \cite{characterOptimizingInference}. MixAttention combines sliding window attention, where only a small subset of recent tokens is stored in the KV cache, with KV cache sharing across layers. Our experiments demonstrate that MixAttention significantly reduces memory usage and improves inference speed without sacrificing model performance in both short and long-context tasks. We also explore various configurations of this architecture, identifying those that maintain quality across evaluation metrics while optimizing resource efficiency.

\end{abstract}

\section{Introduction}

Transformer-based language models  are getting increasing popular in consumer usage as well as industrial workloads. A general trend seen so far has been that bigger models are better at tasks than smaller models, but that comes at the cost of increased inference cost and slower speed \cite{hoffmann2022training, sardanabeyond}. Further, the memory consumption and latency during inference for causal attention transformer models like Llama \cite{touvron2023llama, dubey2024llama}, GPT \cite{radford2019language}, and Gemini \cite{team2023gemini} increases linearly with the input length. This causes problems for use cases such as Retrieval Augmented Generation (RAG) \cite{lewis2020retrieval}, where the input to the models can become very long \cite{databricksLongContext}.

An important component of the Transformer architecture whose memory footprint grows with model size and input length is its KV cache. When generating the next output token, the transformer model processes all the tokens in its context through the attention mechanism. For causal attention models, the internal representation of the previous tokens in the context is unaffected by the newer tokens, and hence it can be cached. 
This is stored in the KV cache, and its size increase with context length (since it caches information for each token seen so far) and with the size of the model (since there is a separate KV cache for each KV head in the model). 
Larger KV cache not only means more memory consumption by the model, but it also slows down inference because for long inputs, LLM inference can be dominated by the I/O cost of moving the KV cache from HBM to the GPU’s shared memory. Thus, it has become imperative to reduce the size of the KV cache for faster and cost-effective inference with modern LLMs.

Several methods have been proposed for reducing the KV cache size including sparse attention methods \cite{beltagy2020longformer}, reducing the number of KV heads \cite{ainslie2023gqa,shazeer2019fast}, KV quantization \cite{hooper2024kvquant}, inference-time cache sparsification through token eviction \cite{zhang2024h2o}, or even replacing some of the attention layers with State Space Machine (SSM) layers \cite{lieber2024jamba}. 
Most of these methods are compatible with others, for example using GQA with Sliding Window Attention \cite{jiang2023mistral}, using GQA with quantization \cite{hooper2024kvquant, lin2024qserve}, or interleaving SSM layers with Sliding Window Attention layers \cite{lieber2024jamba, ren2024samba}. 
In this paper, we explore such a combination proposed by Character.AI where they combine Sliding Window Attention with KV cache sharing across layers \cite{characterOptimizingInference}. We train and evaluate several variants of this architecture, and find that different ways of combining the two ideas result in very different model abilities. In particular, we find some configurations that match the standard transformer model in most short and long context evals, while being faster and more memory efficient during inference.

\begin{figure}
    \begin{subfigure}{0.4\textwidth}
            \includegraphics[trim={4em 20em 22em 0em},clip,width=\textwidth]{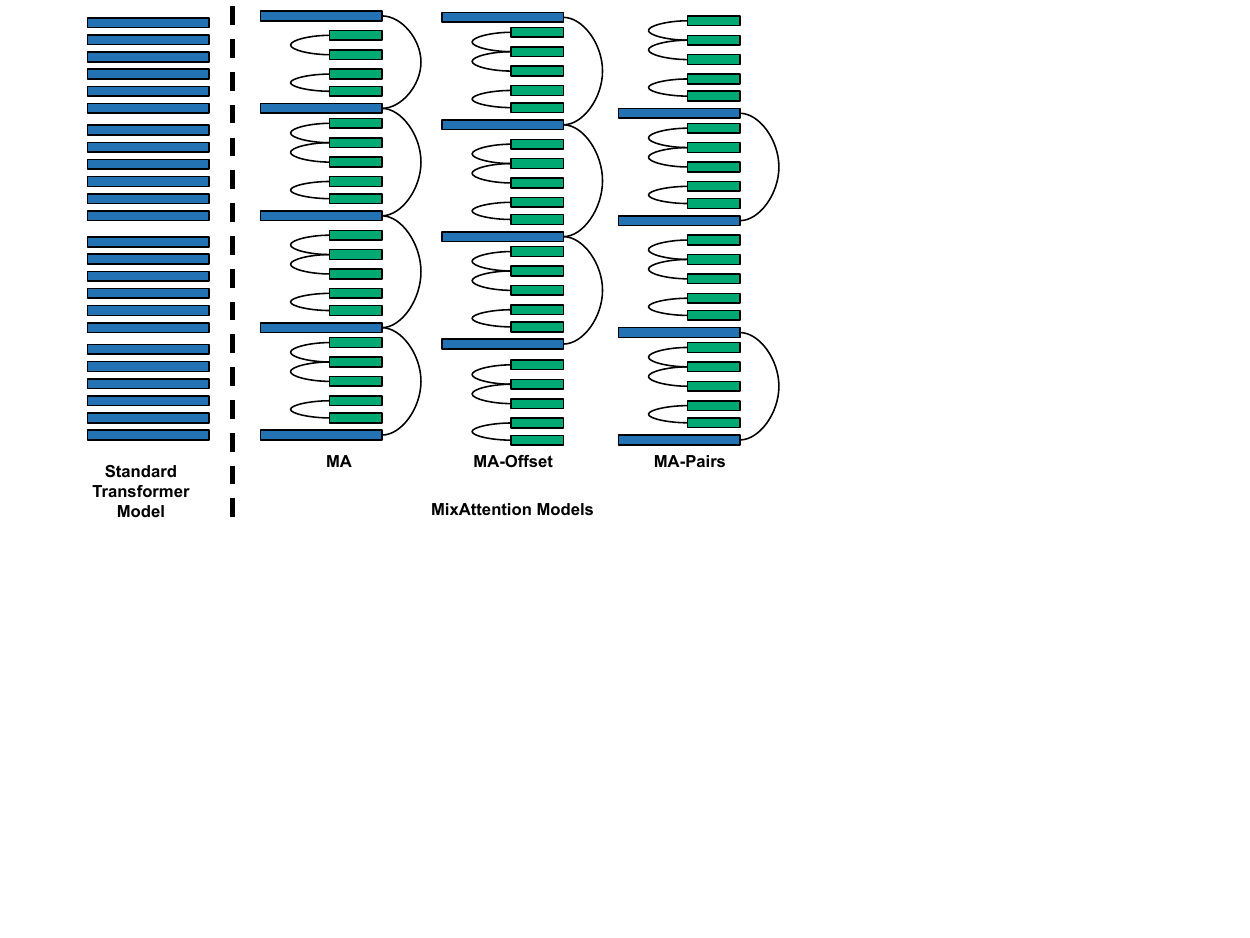}
    \end{subfigure}
    \hspace*{\fill}
    \begin{subfigure}{0.59\textwidth}
        \begin{subfigure}{0.48\textwidth}
        \includegraphics[width=\textwidth]{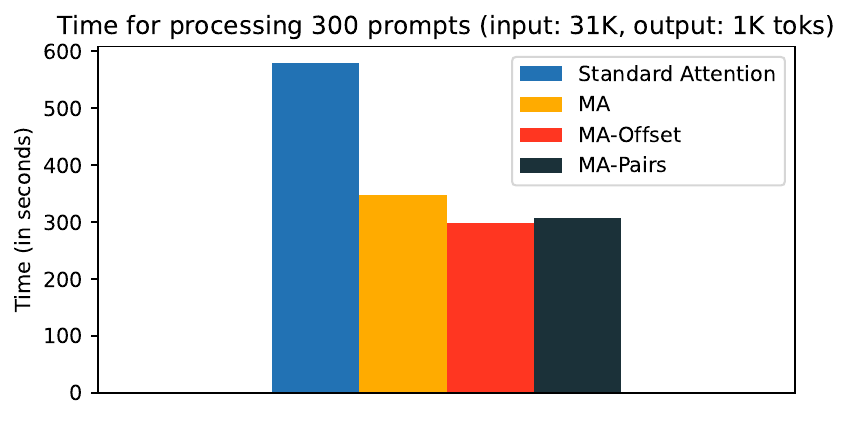}
        \end{subfigure}
        \begin{subfigure}{0.48\textwidth}
        \includegraphics[width=\textwidth]{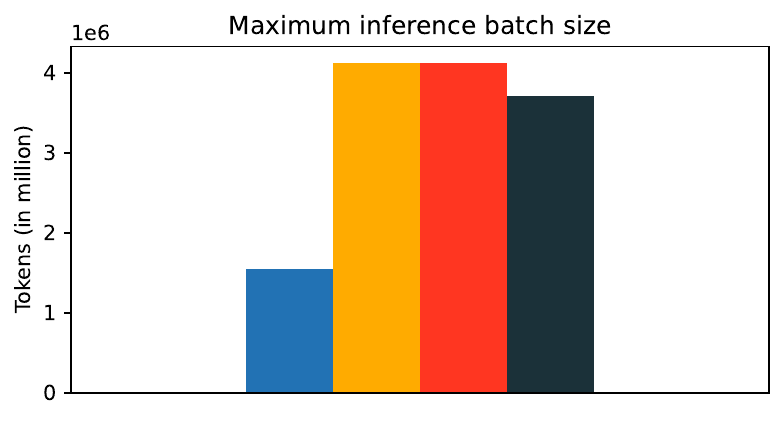}
        \end{subfigure}
        
        \begin{subfigure}{0.48\textwidth}
        \includegraphics[width=\textwidth]{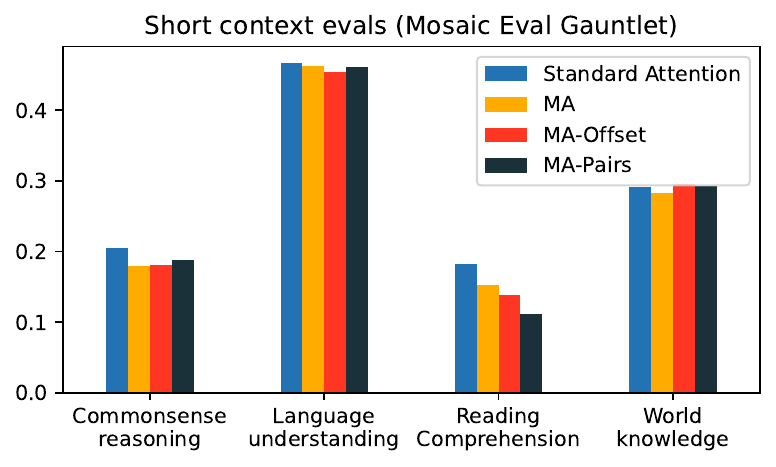}
        \end{subfigure}
        \begin{subfigure}{0.48\textwidth}
        \includegraphics[width=\textwidth]{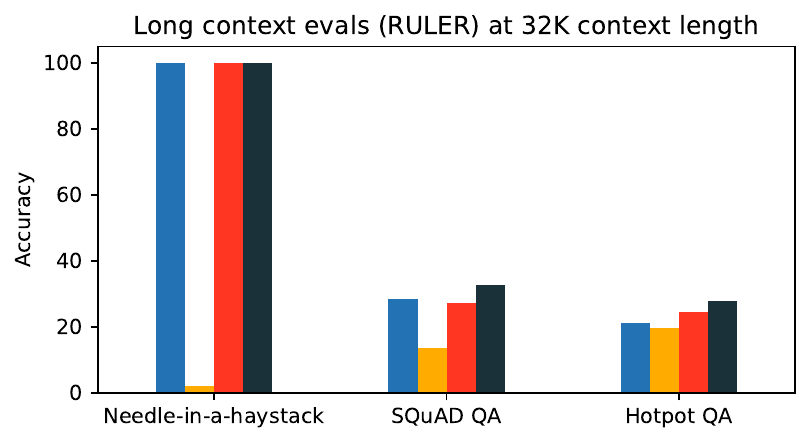}
        \end{subfigure}
    \end{subfigure}
    \caption{(Left) Variants of MixAttention architecture - green bars represent sliding window attention and the curved lines connecting bars represent KV cache sharing. (Right, top row) We see that MixAttention models are faster and use less memory during inference at 32K context length. (Right, bottom row) MixAttention models maintain quality - they match the standard attention model on most evals. The models are all Mixture of Experts with 2B active and 5B total parameters.
}
    \label{fig:intro}
\end{figure}

\subsection{Contributions}
We find that KV cache sharing between layers and adding sliding window layers speeds up inference and reduces inference memory usage while maintaining model quality, although some eval metrics show some degradation (\Cref{fig:intro}). In addition, our ablation experiments show the following:
\begin{itemize}
    \item Having a few standard attention layers is crucial for the model’s long context abilities. In particular, having the standard KV cache computed in the deeper layers is more important for long context abilities than the standard KV cache of the first few layers.
    \item KV cache of standard attention layers can be shared between non-consecutive layers without any observed degradation in long context abilities.
    \item Increasing the KV-cache sharing between sliding window layers too much also hurts the long-context abilities. 
\end{itemize}

\section{Related Work}
Reducing the KV cache size has been an area of active research, with many different approaches. In this section we talk about some of them.
\paragraph{Linear Attention and SSM Models.} Transformer models \cite{vaswani2017attention} differ from traditional Recurrent Neural Networks (RNNs) \cite{sherstinsky2020fundamentals} and modern State Space Models \cite{guefficiently, ren2024samba} in that Transformer models have an internal representation (the KV cache) that grows linearly with the length of the input. This allows RNNs and SSMs to be faster and more memory efficient during inference. However, it has been seen that while such models are competitive with Transformer models on certain tasks, Transformer models still outperform equally-sized pure RNN or pure SSM models on other tasks, especially some long context tasks \cite{waleffe2024empirical}. Thus, hybrid architectures which interleave attention layers and SSM layers have been proposed, that show that such hybrid architectures exhibit good long context abilities \cite{lieber2024jamba, ren2024samba}. Other works have linearized the attention mechanism by replacing the softmax operation with kernelized similarity computation, showing both speed and memory improvements for inference \cite{katharopoulos2020transformers}. 

\paragraph{KV Quantization.} KV quantization works by reducing the precision of the cached key-value (KV) pairs which reduces the overall storage requirements and improves the data movement efficiency during inference \cite{lin2024qserve}. \citet{hooper2024kvquant} combined several novel methods for quantizing the KV cache to achieve significant improvements in long-context language model performance, achieving context lengths up to 10 million tokens while maintaining performance metrics close to those of unquantized models.

\paragraph{KV Eviction.} KV eviction dynamically remove less relevant or older tokens from the KV cache during inference to reduce its size.
This approach ensures that only the most pertinent tokens are retained, helping alleviate memory bottlenecks in long-context tasks. 
\citet{zhang2024h2o} proposed Heavy-Hitter Oracle (H$_2$O), an efficient generative inference approach that selectively evicts tokens from the cache based on their relevance, significantly improving the performance of models operating on large contexts.
\citet{chen2024nacl} introduced NaCl which generalizes H$_2$O and adds random token eviction to retain long context performance while evicting tokens.

\paragraph{KV Head Reduction.} Architectures like Multi-Query Attention (MQA) \cite{shazeer2019fast} and Grouped Query Attention (GQA) \cite{ainslie2023gqa} show that the number of KV heads in the attention layer can be decreased without significantly impacting model performance.
Multi-Query Attention (MQA) simplifies the standard Multi-Head Attention mechanism \cite{vaswani2017attention} by sharing the key and value projections across all attention heads in a layer while retaining independent queries. This approach drastically reduces the size of the KV cache, as fewer unique key-value pairs are stored during inference. However, when serving models on multiple GPUs using Tensor Parallelism (TP) \cite{shoeybi2019megatron}, the single key and value cache must be replicated across the tensor parallel ranks, thus essentially losing a significant fraction of the memory savings gained from using MQA. Hence, Grouped Query Attention (GQA) extends the same idea of query sharing but instead of having one set of keys and values for all the queries, this architecture partitions queries into multiple sets and shares keys and values within each set, where the number of sets (and hence the number of keys and values) often matches the TP rank.

\paragraph{Sparse and Local Attention.}
Sparse and Local Attention mechanisms have been extensively explored as a means to improve the efficiency of Transformer models by reducing the quadratic complexity of traditional global attention. Since these methods focus on attending to only a subset of tokens, they reduce the computational and memory costs during both training and inference.
One of the most prominent methods in this category is Longformer \cite{beltagy2020longformer}, which introduces several variants of local attention mechanism including Sliding Window Attention. Sliding Window Attention and its variants restrict the attention of each token to a fixed window of neighboring tokens, rather than all tokens in the sequence, drastically reducing the number of key-value pairs that need to be stored and processed during inference. This method has been shown to work well but often fails on tasks with long-context dependencies due to the fundamental lack of global attention. Sparse Attention mechanisms further optimize the attention computation by introducing sparsity patterns, where only certain key-value pairs are attended to based on predefined criteria \cite{beltagy2020longformer}. Notably, GPT-3 used interleaving global and local attention layers in its architecture \cite{brown2020language}.

\paragraph{KV Sharing.} KV Sharing \cite{brandon2024reducing, wu2024layer} is a key technique used to reduce the memory footprint of Transformer models during inference by allowing multiple layers to reuse the same key-value (KV) instead of having separate KV pairs for each layer. 
\citet{brandon2024reducing} demonstrate that cross-layer attention, where KV caches are shared across different layers, leads to substantial memory savings without degrading accuracy.

\section{MixAttention}
Standard transformer models use global attention in each layer. To create inference-friendly model architectures, we use a combination of sliding window attention layers, standard attention, and KV cache reuse layers. Below is a brief discussion on each component:

\begin{wrapfigure}[15]{r}{0.5\textwidth}
    \includegraphics[trim={4em 20em 16em 0em},clip,width=0.48\textwidth]{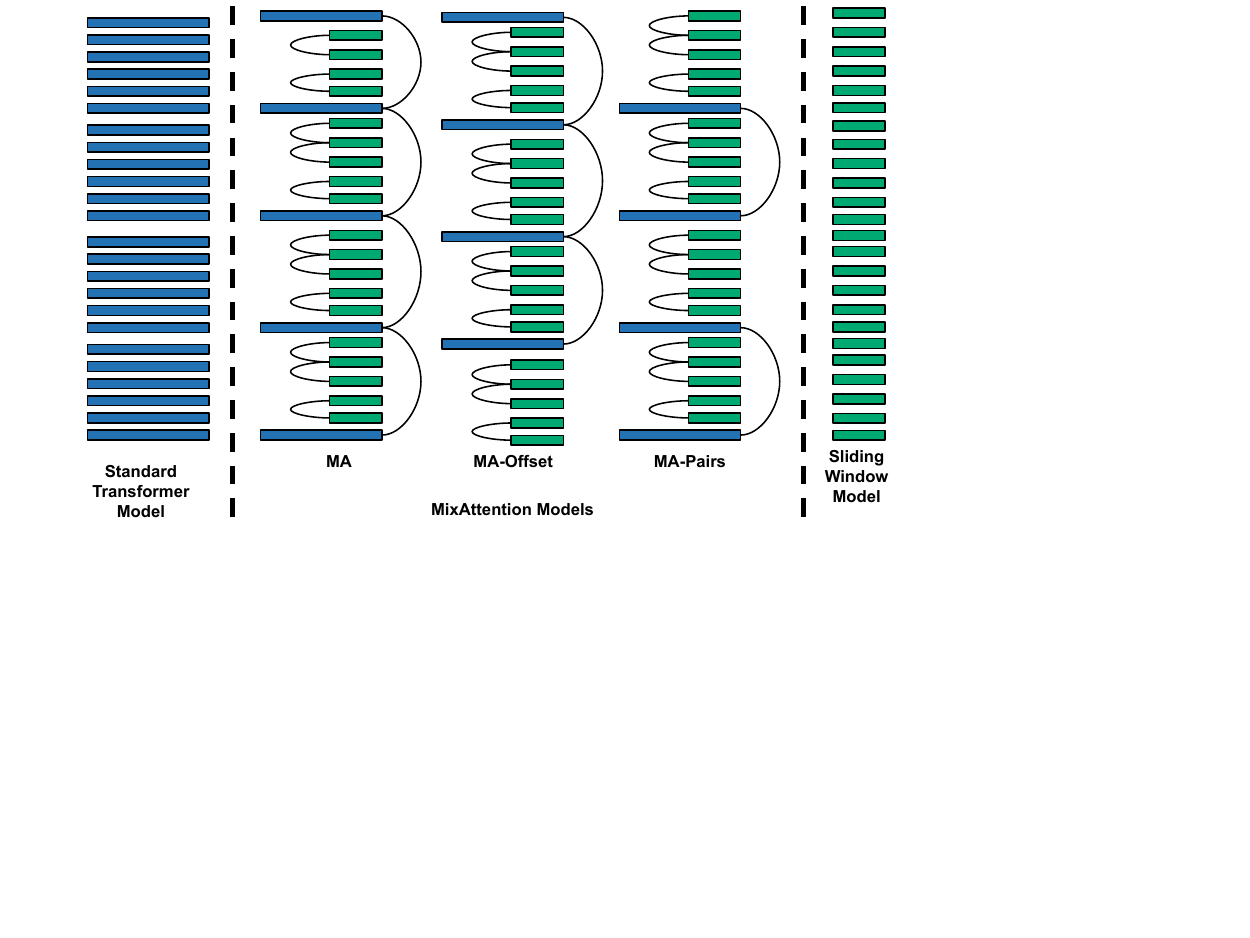}
    \caption{MixAttention: (Left) A standard transformer model where all layers are standard attention layers. (Middle) Inference-friendly models with MixAttention. Green bars represent sliding window attention and the lines connecting bars represent KV cache sharing. (Right) A model where all layers are sliding window attention.}
    \label{fig:mix_attn_main}
\end{wrapfigure}

\paragraph{Sliding Window Attention Layers \cite{beltagy2020longformer}:} In Sliding Window Attention (or Local Attention) with window size $s$, the query only pays attention to the last $s$ keys instead of all the keys preceding it. This means that during inference, the KV cache size needs to only store the KV tensors for the past $s$ tokens instead of storing the KV tensors for all the preceding tokens. In our experiments, we set a window size of $s=1024$ tokens.

\paragraph{Standard Attention Layers:} We found that even though Standard Attention Layers lead to bigger KV caches and slower attention computation compared to Sliding Window Attention, having a few Standard Attention Layers is crucial for the model’s long context abilities.

\paragraph{KV cache reuse \cite{brandon2024reducing,wu2024layer}:} This refers to a layer in the transformer network reusing the KV cache computed by a earlier layer. Hence, if every $l$ layers share KV tensors, then the size of KV cache is reduced by factor of $1/l$.

We experimented with different combinations of the components above to ablate the effects of each of them (\Cref{fig:mix_attn_main}). We found that not only do each of the above components play important roles in long context abilities and inference speed and memory consumption, but also their relative positions and counts have significant effects on those metrics. 

The models we trained are 24-layer Mixture of Experts (MoE) models with 1.64B active and 5.21B total parameters. We used RoPE positional embeddings \cite{su2024roformer}, and increased the RoPE base theta as we increased the context length during training. We used Grouped Query Attention \cite{ainslie2023gqa} with 12 attention heads and 3 KV heads.

\section{Experiments}
\subsection{Training}
We used LLM Foundry \cite{mosaicml2023foundry} to train MixAttention models. Similar to prior work on training long context models  [5, 6], we followed a multi-stage training procedure to impart long context abilities to the models.
\begin{enumerate}
    \item We pretrained the models with a RoPE theta of 0.5M on 101B tokens, where each sequence sequence has been truncated to 4k token length.
   \item To increase the context length, we then trained the model on 9B tokens on a mix of natural language and code data, where the sequences have been truncated to 32k tokens. We increased the RoPE theta to 8M for this stage. When training at 32k context length (\emph{i.e.}, this step and the next step), we trained only the attention weights and froze the rest of the network. We found that this delivered better results than full network training.
   \item Finally, we trained the model on a 32K-length, synthetic, long-context QA dataset [5, 8]. 
   \begin{itemize}
       \item To create the dataset, we took natural language documents and chunked them into 1k-token chunks. Each chunk was then fed to a pretrained instruction model and the model was prompted to generate a question-answer pair based on the chunk. Then, we concatenated chunks from different documents together to serve as the “long context.” At the end of this long context, the question-answer pairs for each of the chunks were added. The loss gradients were computed only on the answer parts of these sequences. 
       \item This phase of training was conducted on 500M tokens (this number includes the tokens from the context, questions, and answers). The RoPE theta was kept at 8M for this stage.
   \end{itemize}

\end{enumerate}
\subsection{Evaluation}
The models were evaluated on the Mosaic Evaluation Gauntlet v 0.3.0 \cite{githubLlmfoundryscriptsevallocal_dataEVAL_GAUNTLETmdMain} to measure model quality across various metrics including reading comprehension, commonsense reasoning, world knowledge, symbolic problem solving, and language understanding. To evaluate the models’ long context abilities, we used RULER \cite{hsieh2024ruler} at a context length of 32000 tokens. RULER is a composite benchmark consisting of 13 individual evals of the following types:
\begin{itemize}
    \item Needle-in-a-haystack (NIAH): These types of evals hide a single or multiple keys and values in a long text, and the model is evaluated on its ability to retrieve the correct value(s) from the long context for a given key(s).
    \item Variable Tracking (VT): This eval provides the model with a long context containing variable assignment statements, and the model is tasked to figure out which variables have a particular value by the end of all the variable assignments.
    \item Common and Frequent Word Extraction (CWE and FWE): These tasks ask the model to extract the most common or frequent words from the text.
    \item Question Answering (QA): Given a long context, the model is asked a question from somewhere in the context and the model is evaluated on whether it can correctly answer that question.
\end{itemize}

We used SGLang \cite{zheng2023efficiently} to deploy our models on 1 NVIDIA H100 GPU to run RULER and get inference speed and memory consumption metrics.

\section{Results}
\subsection{Position and Count of Standard Attention KV Caches}
\begin{wrapfigure}[15]{r}{0.5\textwidth}
\vspace{-1.3em}
    \includegraphics[trim={4em 22em 22em 0em},clip,width=0.48\textwidth]{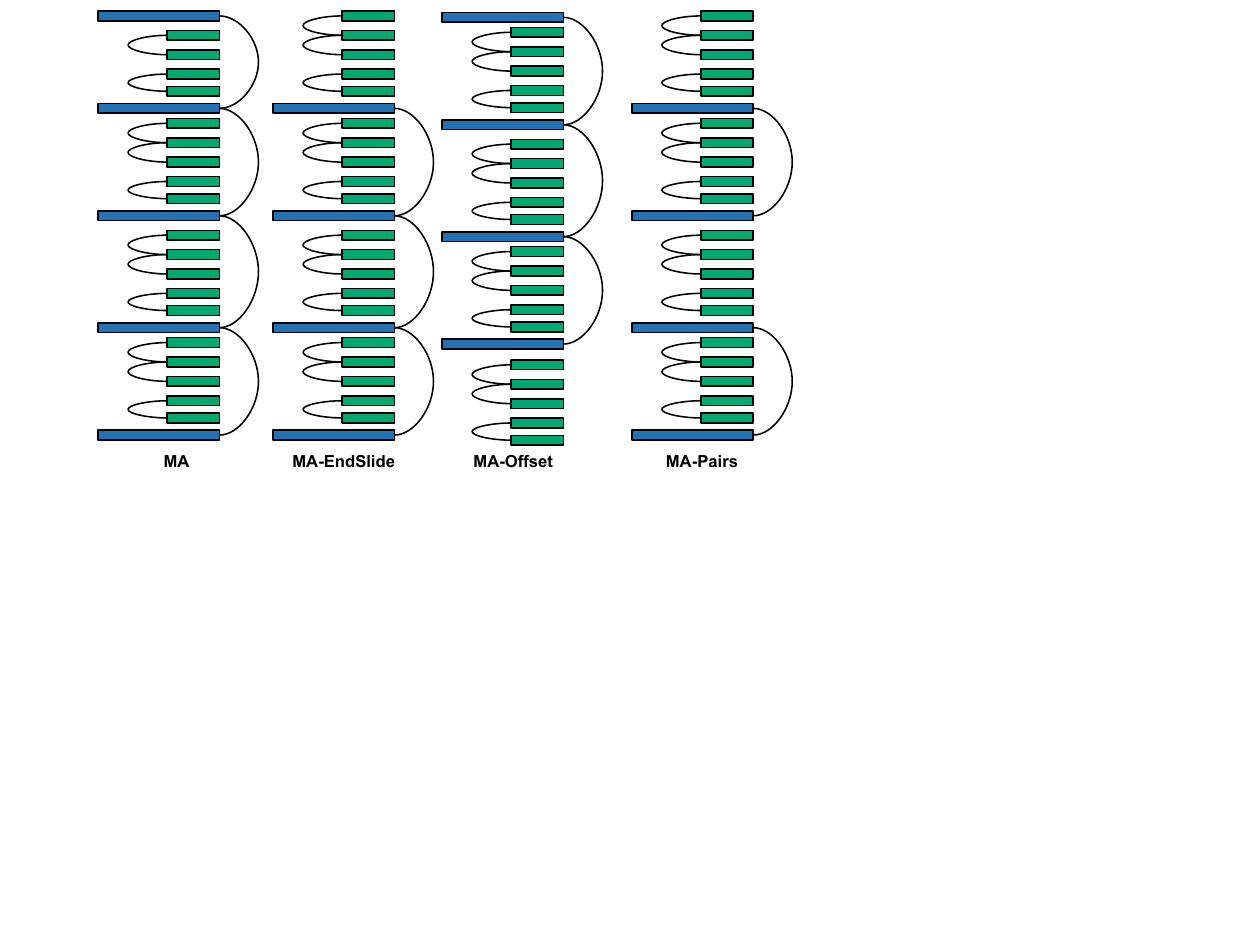}
    \caption{KV Cache position and counts: To measure the effect of the position and count of the standard attention KV caches on MixAttention’s long context abilities, we train and evaluate the 4 models shown above.}
    \label{fig:mix_attn_pos_counts}
\end{wrapfigure}
To measure the effect of the position and count of the standard attention KV caches, we tried four variants (\Cref{fig:mix_attn_pos_counts}). All the configurations are variants of the configuration proposed in Character.AI’s blog \cite{characterOptimizingInference}.

\emph{MA}: This variant has a single standard attention KV cache, which is the KV cache of the first layer. All the other standard attention layers share this KV cache.

\emph{MA-EndSlide}: This variant is the same as MA, but the last layer is a sliding window attention layer. This was done to measure how much having standard attention in the last layer affects long-context abilities.

\emph{MA-Offset}: This variant is similar to MA, but the first standard attention layer is offset to a later layer to allow the model to process the local context for a few layers before the standard attention layer is used to look at longer contexts.

\emph{MA-Pairs}: This variant computes two standard attention KV caches (at the first and thirteenth layer), which are then shared with another standard attention layer each.

We compared these models to a transformer model with Standard Attention and a transformer model with Sliding Window Attention in all layers.

\begin{figure}
    \centering
    \includegraphics[width=\linewidth]{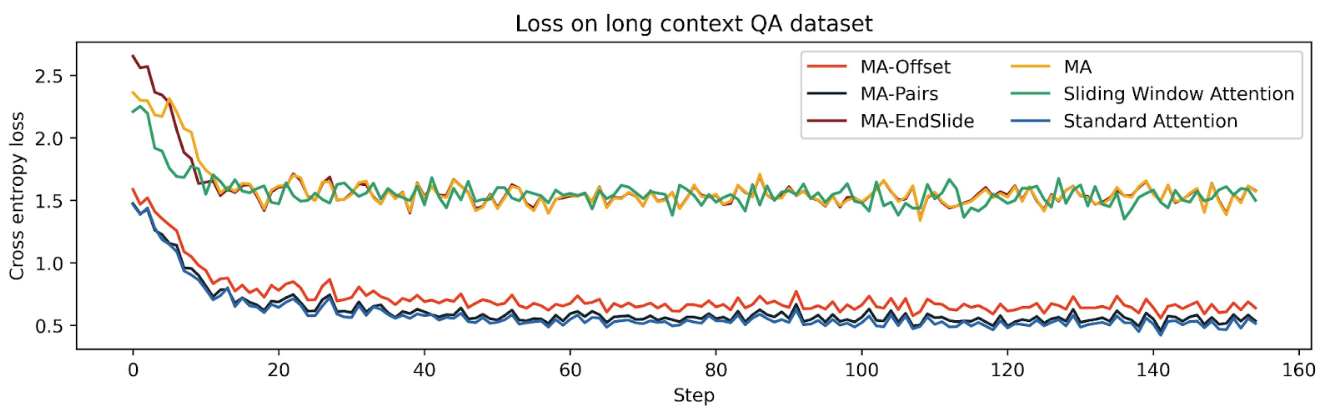}
        
    \includegraphics[width=\linewidth]{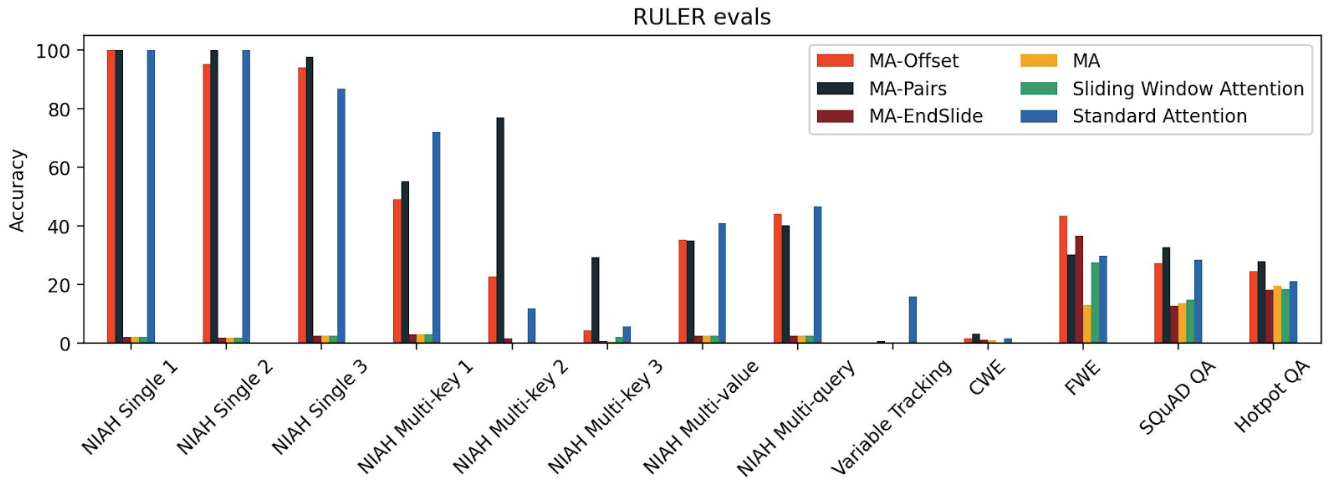}
    \caption{Effect of Standard Attention Layers: (Top) Loss curves of the models when fine tuning on long context QA dataset. (Bottom) RULER evals for the models. MA and MA-EndSlide perform poorly on long context tasks whereas MA-Offset and MA-Pairs perform well. This indicates that having a standard attention KV cache which is computed in later layers is important for long context abilities. We also found that the loss on long context QA dataset correlates well with the model’s long context abilities.}
    \label{fig:results-pos-counts}
\end{figure}

While the loss curves in Stages 1 and 2 of training were close for all the models, we found that in Stage 3 (training on long context QA dataset), there was a clear bifurcation in the loss curves (\Cref{fig:results-pos-counts}, top). In particular, we see that configurations MA and MA-EndSlide show much worse loss than the others. These results are consistent with the long context RULER evals, where we found that MA and MA-EndSlide performed much worse than others (\Cref{fig:results-pos-counts}, bottom). Their performance was similar to the performance of the network with only sliding window attention in all layers. We think the loss in Stage 3 correlates well with RULER evals because unlike Stages 1 and 2, which were next-word prediction tasks where local context was sufficient to predict the next word most of the time, in Stage 3 the model needed to retrieve the correct information from potentially long-distance context to answer the questions. 
As we see from the RULER evals, MA-Offset and MA-Pairs have better long-context abilities than MA and MA-EndSlide across all the categories. 
Both MA and MA-EndSlide have only one standard attention KV-cache, which is computed in the first layer, 
whereas both MA-Offset and MA-Pairs have at least one standard attention KV-cache which is computed in deeper layers. 
Hence, this indicates that having at least one standard attention KV cache that is computed in the deeper layers of a transformer model is necessary for good long-context abilities.

\subsection{KV~cache~sharing~in~sliding~window~layers}
\begin{wrapfigure}[13]{r}{0.5\textwidth}
\vspace{-2.6em}
    \includegraphics[trim={22em 22em 5em 0em},clip,width=0.48\textwidth]{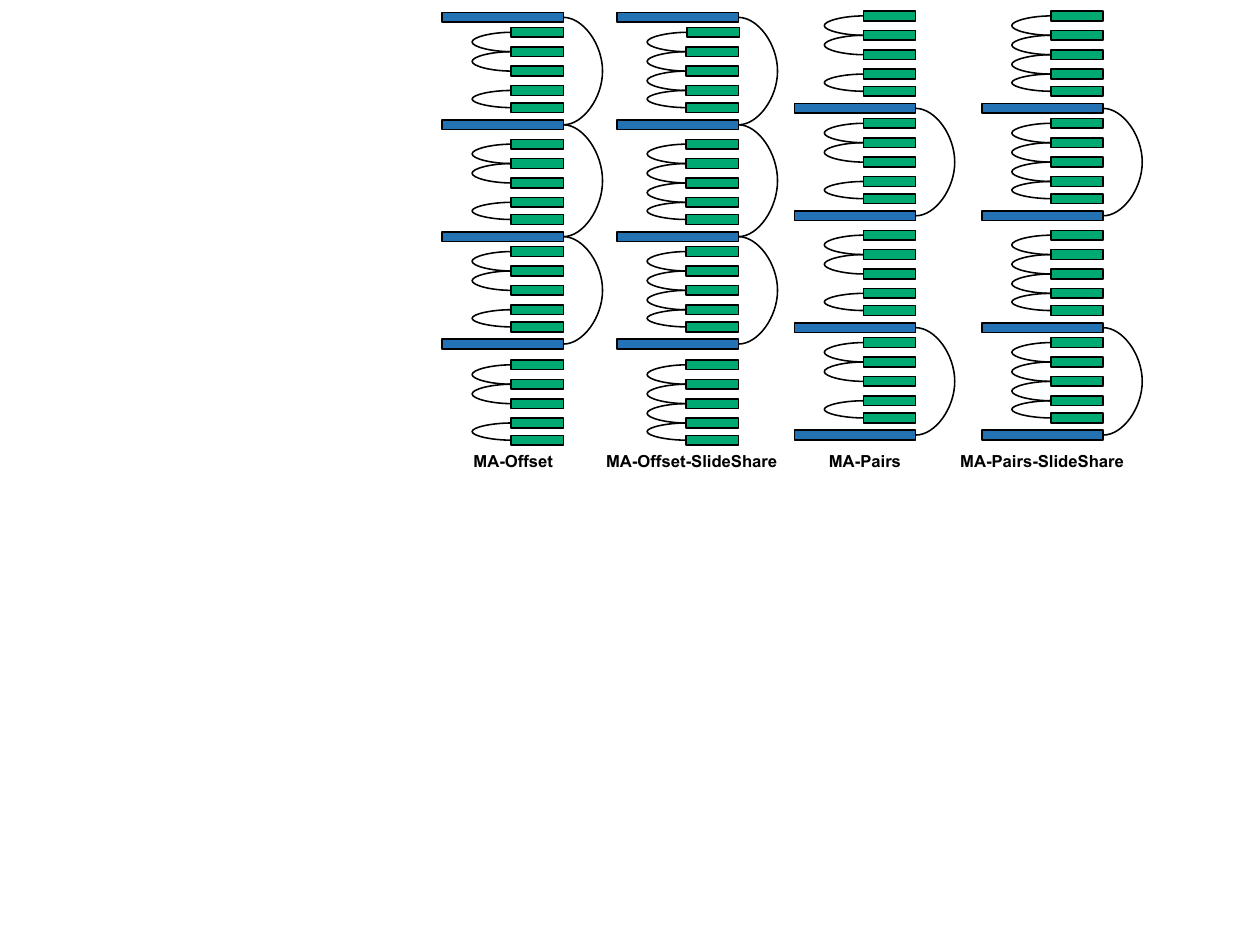}
    \caption{Increasing KV cache sharing in sliding window layers: To measure the effect of KV cache sharing in the sliding window layers, we compared the architectures shown in the figure above.}
    \label{fig:mix_attn_slideshare}
\end{wrapfigure}
\begin{figure}
    \centering
    \includegraphics[width=\linewidth]{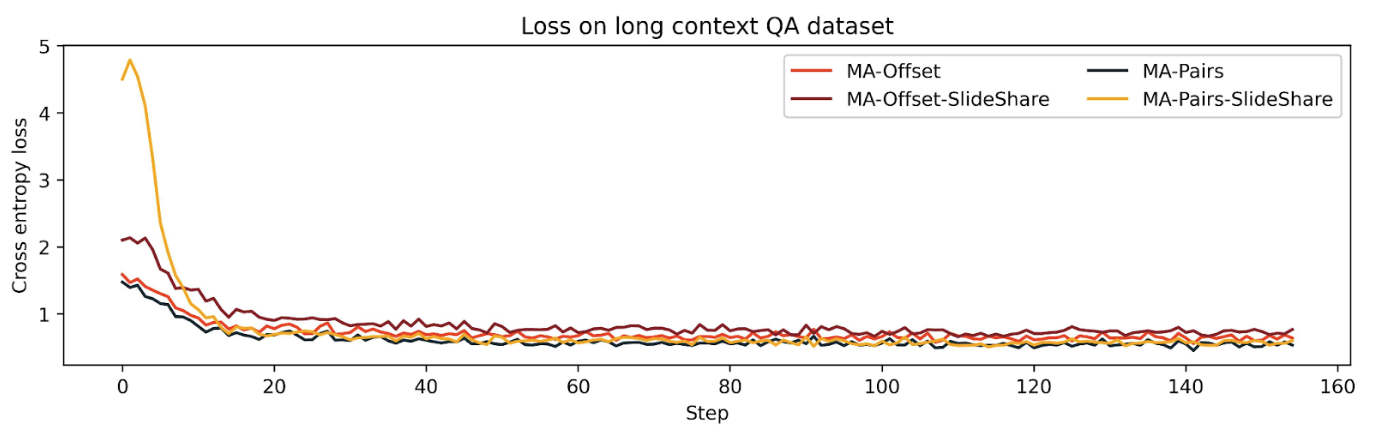}
        
    \includegraphics[width=\linewidth]{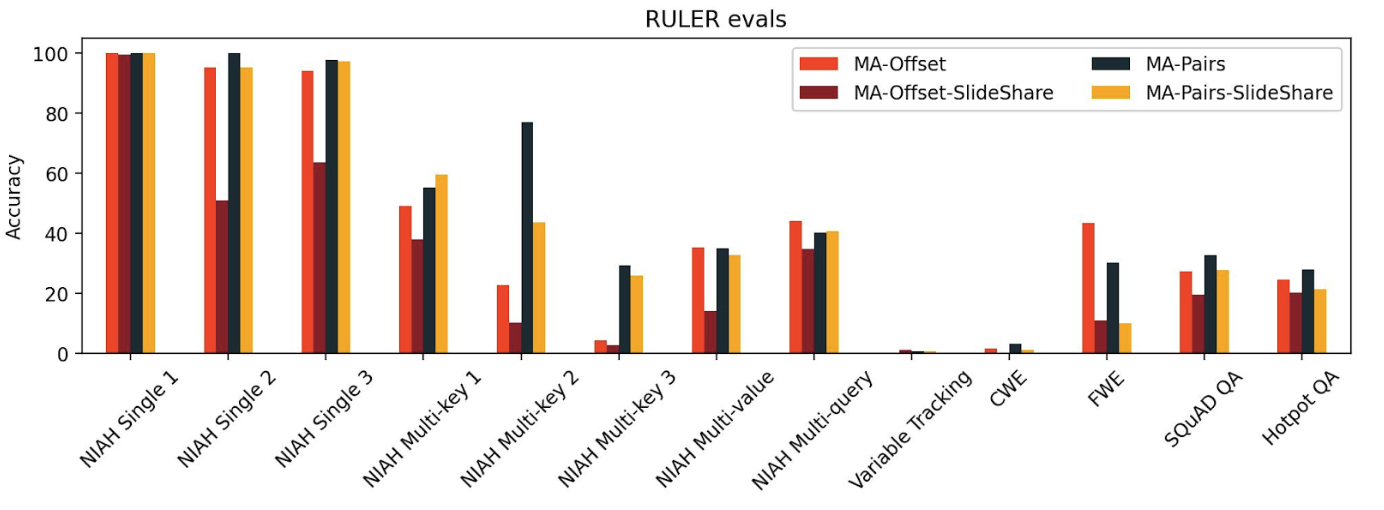}
    \caption{Effect of increasing KV cache sharing in sliding window layers: (Top) Loss curves of the models when fine tuning on long context QA dataset. (Bottom) RULER evals for the models. We found that increasing the KV cache sharing in sliding window layers worsened long context abilities of MixAttention Models.}
    \label{fig:results-slideshare}
\end{figure}

We found that increasing the sharing between sliding window layers (\Cref{fig:mix_attn_slideshare}) degraded the model’s long context performance: MA-Offset-SlideShare was worse than MA-Offset and MA-Pairs-SlideShare was worse than MA-Pairs (\Cref{fig:results-slideshare}). This shows that the KV cache sharing pattern amongst the sliding window layers is also important for long context abilities.
We have provided some more ablation experiments in the appendix.

\subsection{Gauntlet Evals}

\begin{figure}
\vspace{-2.5em}
    \centering
    \includegraphics[width=\linewidth]{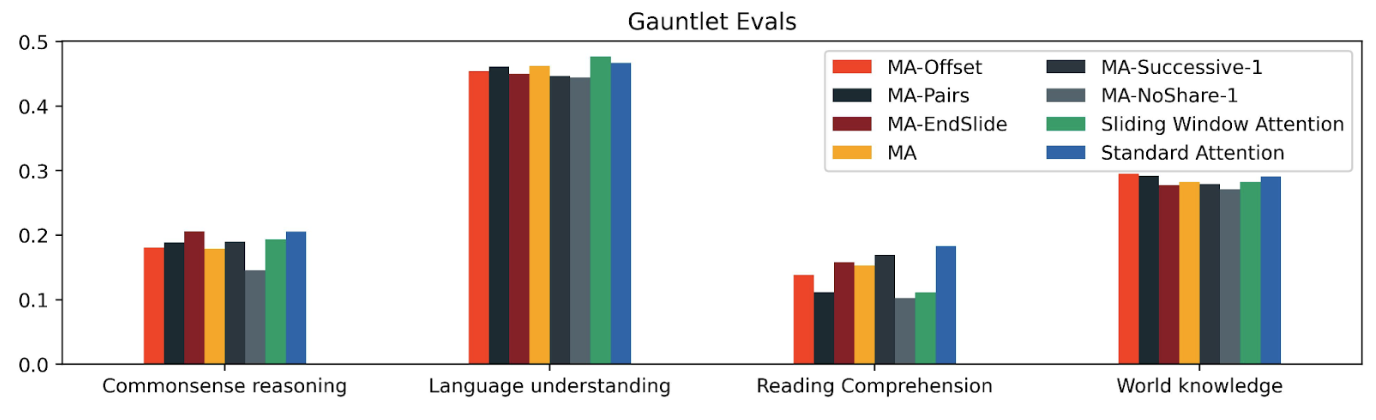}
    \caption{Performance of MixAttention models on the Eval Gauntlet: We found that MixAttention models have similar eval metrics to the baseline model on commonsense reasoning, language understanding, and world knowledge. However, we see that they perform worse on reading comprehension.}
    \label{fig:gauntlet-main}
\end{figure}

Using the Mosaic Eval Gauntlet v0.3.0 \cite{githubLlmfoundryscriptsevallocal_dataEVAL_GAUNTLETmdMain}, we measured the performance of MixAttention models on standard tasks like MMLU \cite{hendrycksmeasuring}, HellaSwag \cite{zellers2019hellaswag}, etc. to verify that they retain good shorter context abilities. All of the tasks in this eval suite have context lengths of less than a few thousand tokens.

We found that MixAttention models have similar eval metrics to the baseline model on commonsense reasoning, language understanding, and world knowledge. However, we see that they perform worse on reading comprehension. An interesting open question is if a different MixAttention configuration or training MixAttention models longer can recover the reading comprehension abilities.

\subsection{Inference Speed and Memory Consumption}

\begin{figure}
    \centering
    \includegraphics[width=\linewidth]{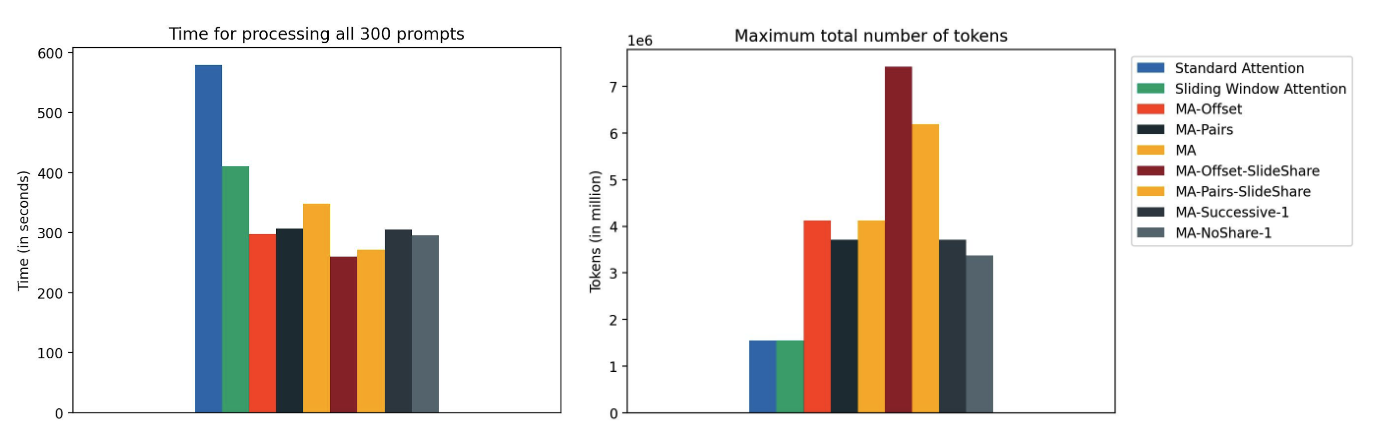}
    \caption{Inference with MixAttention: (Left) MixAttention models have significantly faster inference than standard transformers. (Right) MixAttention models can support more tokens, and thus larger batch sizes, during inference.}
    \label{fig:inference-speed-mem}
\end{figure}

We benchmarked the inference speed and memory consumption of MixAttention models by deploying them on a single NVIDIA H100 GPU using SGLang and querying them with 300 prompts, with input length 31000 and output length 1000. In \Cref{fig:inference-speed-mem}, we see that the inference speed of MixAttention models is much faster than standard attention models. We also see in \Cref{fig:inference-speed-mem} that with MixAttention, we can support a much larger inference batch size in terms of total number of tokens. 

Note that the implementation of Sliding Window Attention in SGLang at the time of writing this paper did not optimize the memory consumption for sliding window attention; hence in \Cref{fig:inference-speed-mem}, sliding window attention has the same maximum number of tokens as the standard attention model. Optimizing the memory consumption for sliding window attention should further increase the maximum number of tokens that MixAttention can support during inference.

\section{Conclusion}
We find that MixAttention models are competitive with standard attention models on both long- and short-context abilities while being faster during inference and supporting larger batch sizes. We note that on some long context tasks like Variable Tracking and Common Word Extraction, neither MixAttention nor standard attention models perform well. We believe this was because our models weren’t trained long enough or the models need a different kind of long context data to be trained for such tasks. More research needs to be done to measure the impact of MixAttention architectures on such metrics.

We encourage others to explore more MixAttention architectures to learn more about them. Below are a few observations to help with further research:

\begin{itemize}
    \item Adding a standard attention layer in the initial layers by itself does not seem to help long context abilities (for example, see MA-NoShare-1 in the appendix), even if the KV cache from that layer is reused in layers deeper into the network (MA and MA-EndSlide). Hence we recommend placing the first standard attention layer deeper in the network (like MA-Offset) or having multiple standard attention layers, at least one of which is computed at a deeper layer (like MA-Pairs).
    \item Sliding window layers also contribute to the model’s long context abilities. Increasing the KV cache sharing amongst the sliding window layers worsened long context abilities (MA-Offset-SlideShare and MA-Pairs-SlideShare). For that reason, we think that the 2-3 sharing pattern in sliding window layers \cite{characterOptimizingInference} seems to strike a good balance.
    \item Sharing standard attention KV caches between consecutive layers gave mixed results, with slightly worse accuracy on long context QA tasks (see the appendix).
    \item In our experiments, MA-Offset and MA-Pair showed great speedup and memory savings during inference, while also maintaining long and short context abilities. Hence, MA-Offset and MA-Pairs might be good configurations for further research.
\end{itemize} 

In general, there is a large hyperparameter space to explore, and we look forward to seeing a variety of new strategies for reducing the cost of inference via combinations of sliding window attention and KV cache reuse.

\section*{Acknowledgments}
The authors thank Alex Trott, Davis Blalock, Emily Hutson, Jacob Portes, and Jonathan Frankle for their helpful feedback on this paper.

\bibliographystyle{plainnat}
\bibliography{ref.bib}
\newpage
\appendix

\section{Additional Ablations}
\subsection{Sharing Standard Attention KV caches between consecutive layers}

\begin{wrapfigure}[13]{r}{0.5\textwidth}
    \centering
    \vspace{-1.2em}
    \includegraphics[width=0.5\textwidth]{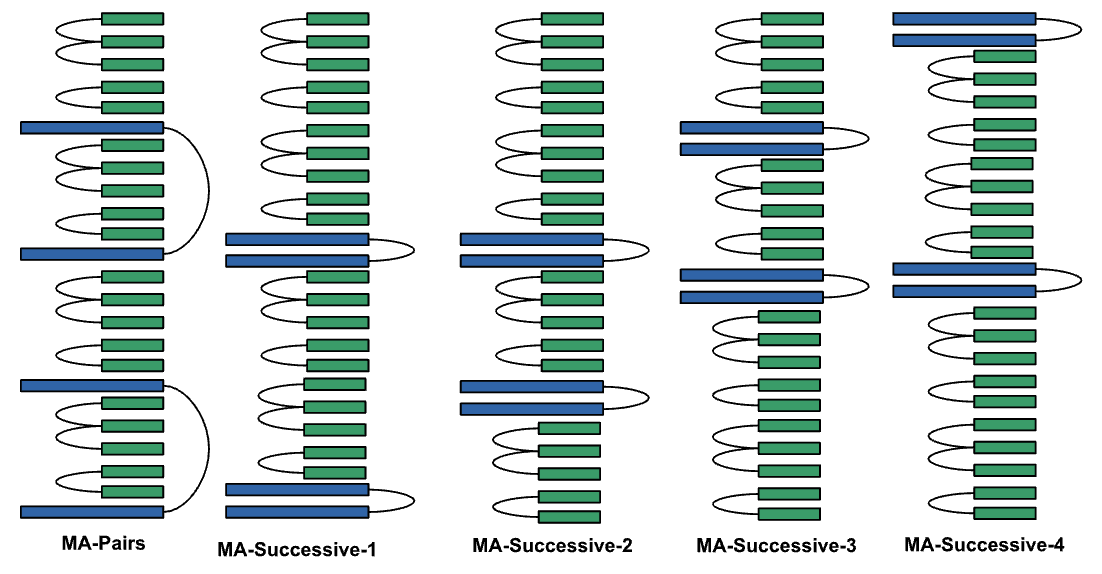}
    \caption{KV cache sharing between consecutive layers: To measure the effect of KV cache sharing between consecutive layers, we tried the four configurations above.}
    \label{fig:mix_attn_successive}
\end{wrapfigure}
Since the transformer layers progressively update the latent representation of a token as it progresses through the layers, the Query, Key, and Value tensors might have significantly different representations for layers that are far apart. Hence, it might make more sense to share KV caches between consecutive layers. To test this, we compared four such configurations: MA-Successive-1, MA-Successive-2, MA-Successive-3, and MA-Successive-4 against MA-Pairs. These configurations vary the positions of the standard KV attention layers and the distance between the consecutive pairs of standard KV attention layers.

\begin{figure}
    \centering
    \includegraphics[width=\linewidth]{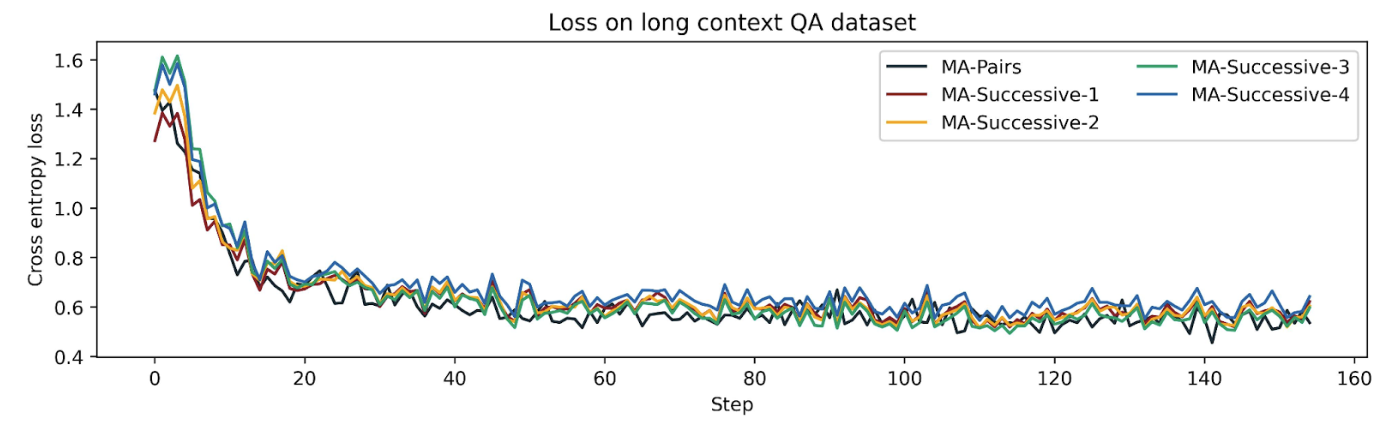}
        
    \includegraphics[width=\linewidth]{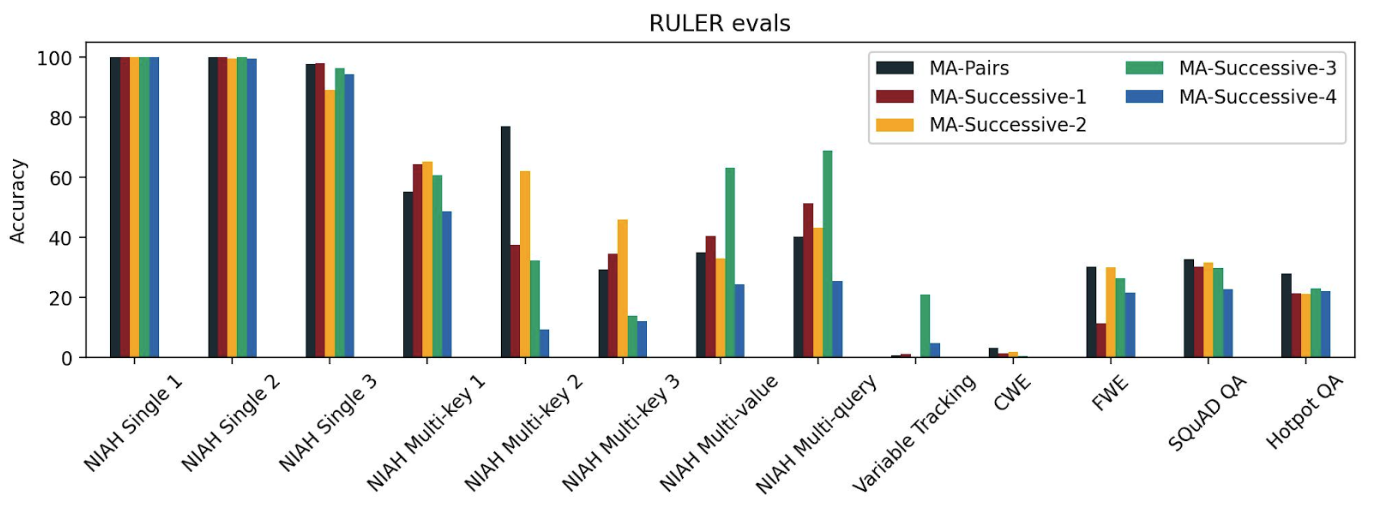}
    \caption{Effect of KV cache sharing between consecutive layers: (Top) Loss curves of the models when fine tuning on long context QA dataset. (Bottom) RULER evals for the models. We found that KV cache sharing between consecutive layers does not consistently increase long context abilities across all evals. However, for tasks like  SQuAD QA and Hotpot QA, which can be indicative of long context RAG abilities, the performance was slightly worse when sharing KV cache between consecutive layers.}
    \label{fig:results-successive}
\end{figure}
We determined that all the models have similar loss curves and similar performance on NIAH single 1, 2, and 3 tasks, which we consider to be the easiest long context tasks. However, we did not see a consistent pattern across the other NIAH tasks. For long context QA tasks, we found that MA-Pairs was slightly better than the others. These results indicate that sharing standard attention KV cache between layers that are further apart does not lead to any significant degradation in long context abilities as compared to sharing standard attention KV cache between consecutive layers.

\subsection{Effect of sharing standard attention KV cache}
\begin{figure}
    \centering
    \includegraphics[width=0.5\textwidth]{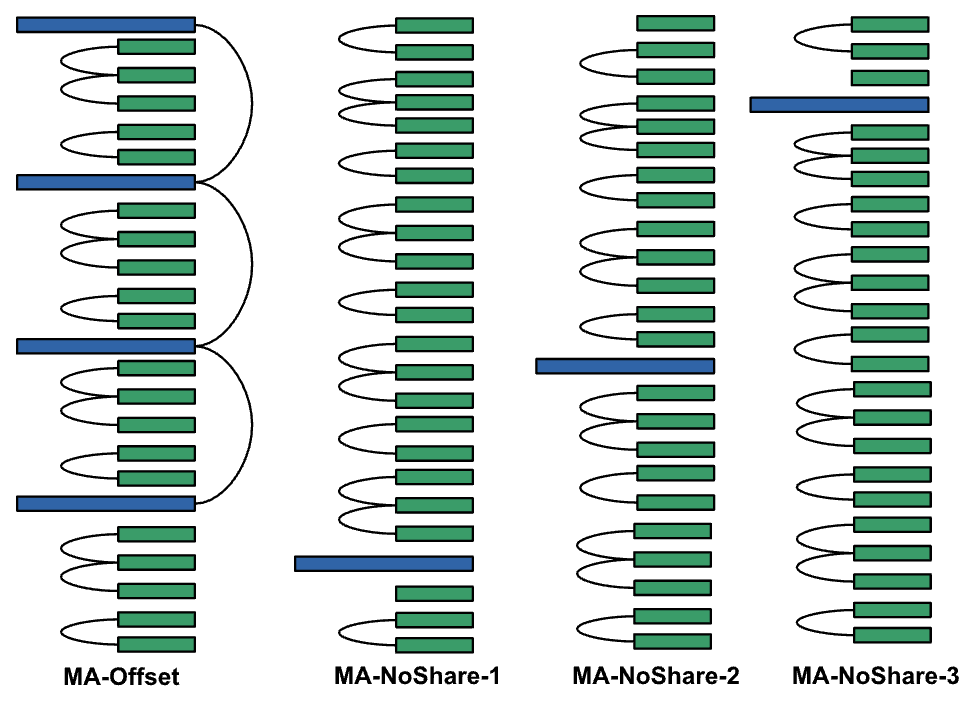}
    \caption{KV cache sharing between consecutive layers: To measure the effect of KV cache sharing between consecutive layers, we tried the four configurations above.}
    \label{fig:mix_attn_noshare}
\end{figure}

To test the effect of sharing KV cache between standard attention layers, we tried out three configurations: MA-NoShare-1, MA-NoShare-2, and MA-NoShare-3 (\Cref{fig:mix_attn_noshare}). We found that MA-NoShare-1 performed very badly on RULER, indicating its lack of long context abilities. However, MA-NoShare-2 and MA-NoShare-3 were comparable to MA-Offset on long context tasks. Hence, we think that further research is needed to ascertain the effects of sharing standard attention KV cache.

\begin{figure}
    \centering
    \includegraphics[width=\linewidth]{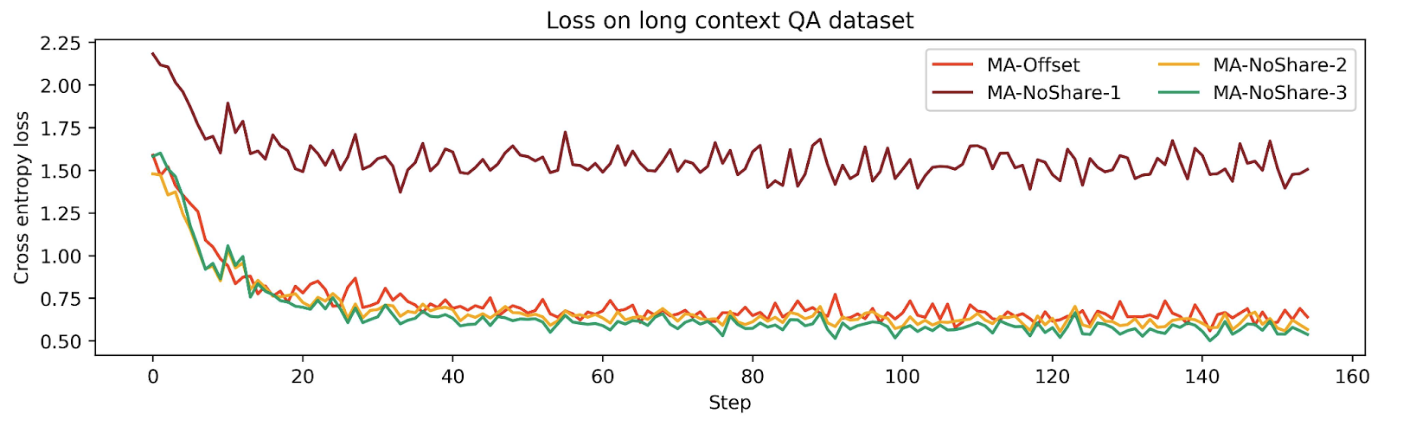}
        
    \includegraphics[width=\linewidth]{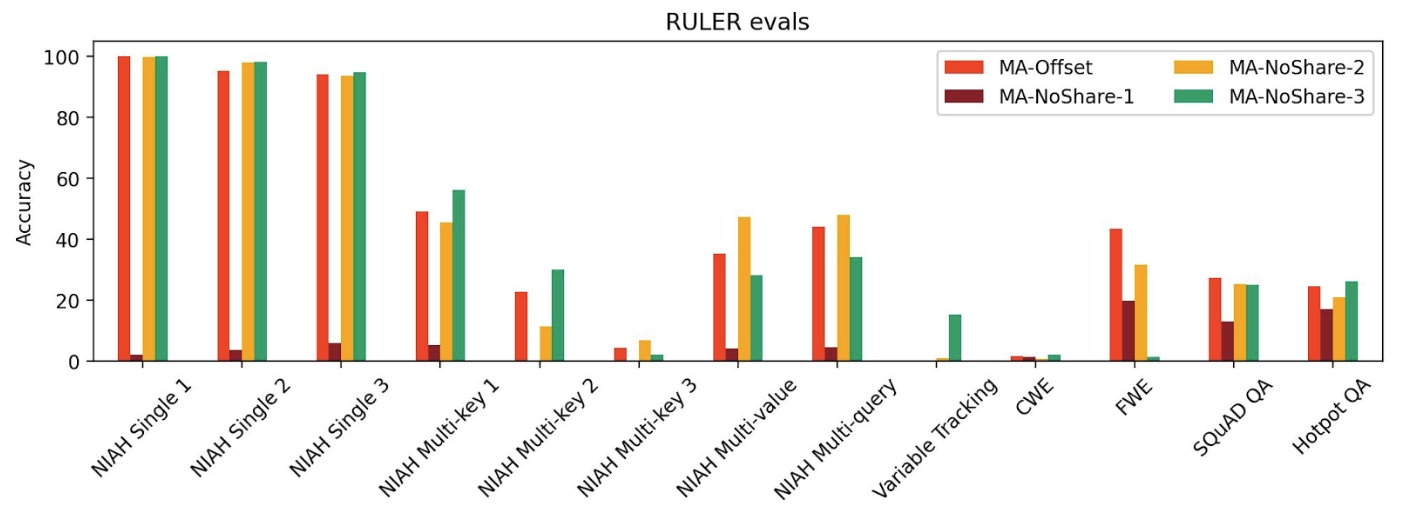}
    \caption{Effect of no standard attention KV-cache sharing: (Top) Loss curves of the models when fine tuning on long context QA dataset. (Bottom) RULER evals for the models. We found that both MA-NoShare-2 and MA-NoShare-3 were comparable with MA-Offset.}
    \label{fig:results-noshare}
\end{figure}
\end{document}